\documentclass[11pt,a4paper]{article}
\usepackage[hyperref]{conll2018}
\usepackage{times}
\usepackage{latexsym}
\usepackage{amsmath}
\usepackage{framed}
\usepackage{booktabs}
\usepackage{amssymb}
\usepackage[colorinlistoftodos,prependcaption,textsize=tiny]{todonotes}
\usepackage[font=small]{caption}

\usepackage{url}
  
  \newcommand{\quickfigA}[4]{  
  \begin{figure*}
    \begin{minipage}{\linewidth}
      \makebox[\linewidth]{
        \includegraphics[keepaspectratio=true,scale=#3]{#1}}
      \captionof{figure}{#2}
      \label{#4}
    \end{minipage} 
  \end{figure*}
  }

\usepackage{url}

\aclfinalcopy 

\title{Unsupervised Sentence Compression using Denoising Auto-Encoders}

 \author{Thibault~Fevry$^*$  \\
    Center for Data Science \\
    New York University \\
    {\tt Thibault.Fevry@nyu.edu} \\\And
    Jason Phang$^*$ \\
    Center for Data Science \\
    New York University \\
    {\tt jasonphang@nyu.edu} \\}

\date{}

\begin{document}
\maketitle
{\let\thefootnote\relax\footnote{{$^*$ Denotes equal contribution}}}

\begin{abstract}
In sentence compression, the task of shortening sentences while retaining the original meaning, models tend to be trained on large corpora containing pairs of verbose and compressed sentences. To remove the need for paired corpora, we emulate a summarization task and add noise to extend sentences and train a denoising auto-encoder to recover the original, constructing an end-to-end training regime without the need for any examples of compressed sentences. We conduct a human evaluation of our model on a standard text summarization dataset and show that it performs comparably to a supervised baseline based on grammatical correctness and retention of meaning. Despite being exposed to no target data, our unsupervised models learn to generate imperfect but reasonably readable sentence summaries. Although we underperform supervised models based on ROUGE scores, our models are competitive with a supervised baseline based on human evaluation for grammatical correctness and retention of meaning.\\

\end{abstract}

\section{Introduction}

Sentence compression is the task of condensing a longer sentence into a shorter one that still retains the meaning of the original. Past models for sentence compression have tended to rely heavily on strong linguistic priors such as syntactic rules or heuristics \cite{dorr2003hedge, cohn2008sentence}. More recent work using deep learning involves models trained without strong linguistic priors, instead requiring large corpora consisting of pairs of longer and shorter sentences \cite{lang_latent}.

Sentence compression can also be can be seen as a ``scaled down version of the text summarization problem" \cite{knight2002summarization}. Within text summarization, two broad approaches exist: \textit{extractive} approaches extract explicit tokens or phrases from the reference text, whereas \textit{abstractive} approaches involve a compressed paraphrasing of the reference text, similar to the approach humans might take \cite{jing2000sentence, jing2002using}. 

In the related domain of machine translation, a task that also involves learning a mapping from one string of tokens to another, state of the art models using deep learning techniques are trained on large parallel corpora. Recent promising work on unsupervised neural machine translation \cite{artetxe2017unsupervised,lample2017unsupervised} has shown that with the right training regime, it is possible to train models for machine translation between two languages given only two unpaired monolingual corpora.

In this paper, we apply neural text summarization techniques to the task of sentence compression, focusing on on extractive summarization. However, we depart significantly from prior work by taking a fully unsupervised training approach. Beyond not using parallel corpora, we train our model using a single corpus. In contrast to unsupervised neural machine translation, which still uses two corpora, we do not have separate corpora of longer and shorter sentences.

We show that a simple denoising auto-encoder model, trained on removing and reordering words from a noised input sequence, can learn effective sentence compression, generating shorter sequences of reasonably grammatical text that retain the original meaning. While the models are still prone to both errors in grammar and meaning, we believe that this is a strong step toward reducing reliance on paired corpora.

We evaluate our model using both a standard text-summarization benchmark as well as human evaluation of compressed sentences based on grammatical correctness and retention of meaning. Although our models do not capture the written style of the target summaries (headlines), they still produce reasonably readable and accurate compressed sentence summaries, without ever being exposed to any target sentence summaries. We find that our model underperforms based on ROUGE metrics, especially compared to supervised models, but performs competitively with supervised baselines in human evaluation. We further show that providing the model with a sentence embedding of the original sentence leads to better ROUGE scores but worse human evaluation scores. However, both unsupervised and supervised methods still fall short based on human evaluation, and effective sentence compression and summarization remains an open problem.

\section{Related work}

Early sentence compression approaches were extractive, focusing on deletion of uninformative words from sentences through learned rules \cite{knight2002summarization} or linguistically-motivated heuristics \cite{dorr2003hedge}. The first abstractive approaches also relied on learned syntactic transformations \cite{cohn2008sentence}.

Recent work in automated text summarization has seen the application of sequence-to-sequence models to automatic summarization, including both extractive \cite{nallapati2017summarunner} and abstractive \cite{rush2015neural, chopra2016abstractive, nallapati2016abstractive, paulus2017deep,fan2017controllable} approaches, as well as hybrids of both \cite{see2017get}. Although these methods have achieved state-of-the-art results, they are constrained by their need for large amounts paired document-summary data.

\citet{lang_latent} seek to overcome this shortcoming by training separate compressor and reconstruction models, allowing for training based on both paired (supervised) and unlabeled (unsupervised) data. For their compressor, they train a discrete variational auto-encoder for sentence compression and use the REINFORCE algorithm to allow end-to-end training. They further use a pre-trained language model as a prior for their compression model to induce their compressed output to be grammatical. However, their reported results are still based on models trained on at least 500k instances of paired data.

In machine translation, unsupervised methods for aligning word embeddings using only unmatched bilingual corpora, trained with only small seed dictionaries, \cite{mikolov2013exploiting, lazaridou2015hubness}, adversarial training on similar corpora \cite{zhang2017adversarial, conneau2017word} or even on distant corpora and languages \cite{robustbilingual} have enabled the development of unsupervised machine translation \cite{artetxe2017unsupervised, lample2017unsupervised}. However, it is not clear how to adapt these methods for summarization where the task is to shorten the reference rather than translate it. \citet{wang2018learning} train a generative adversarial network to encode references into a latent space and decode them in summaries using only unmatched document-summary pairs. However, in contrast with machine translation where monolingual data is plentiful and paired data scarce, summaries are paired with their respective documents when they exist, thus limiting the usefulness of such approaches. In contrast, our method requires no summary corpora.

Denoising auto-encoders \cite{vincent2008extracting} have been successfully used in natural language processing for building sentence embeddings \cite{hill2016learning}, training unsupervised translation models \cite{artetxe2017unsupervised} or for natural language generation in narrow domains \cite{freitag2018unsupervised}. In all those instances, the added noise takes the form of random deletion of words and word swapping or shuffling. Although our noising mechanism relies on adding rather than removing words, we take some inspiration from these works.

Work in sentence simplification (see \citet{shardlow2014survey} for a survey) has some similarities with sentence compression, but it differs in that the key focus is on making sentences more easily understandable rather than shorter. Though word deletion is used, sentence simplification methods feature sentence splitting and word simplification which are not usually present in sentence compression. Furthermore, these methods often rely heavily on learned rules (e.g lexical simplification as in \citet{biran2011putting}), integer linear programming and sentence parse trees which makes them starkly different from our deep learning-based approach. The exceptions that adopt end-to-end approaches, such as \citet{filippova2015sentence}, are usually supervised and focus on word deletion.

  \section{Methods}

  \subsection{Model}
  
    Our core model is based on a standard attentional encoder-decoder \cite{DBLP:journals/corr/BahdanauCB14}, consisting of multiple layers bi-directional long short-term memory networks in both the encoder and decoder, with negative-log likelihood as our loss function. We detail below the training regime and model modifications to apply the denoising auto-encoding paradigm to sentence compression.

  \subsection{Additive Noising}
    Since we do not use paired sentence compression data with which to train our model in a supervised way, we simulate a supervised training regime by modifying a denoising auto-encoder (DAE) training regime to more closely resemble supervised sentence compression. 
    Given a reference sentence, we extend and shuffle the input sentence, and then train our model to recover the original reference sentence. In doing so, the model has to exclude and reorder words, and hence learns to output shorter but grammatically correct sentences. 
    
    \quickfigA{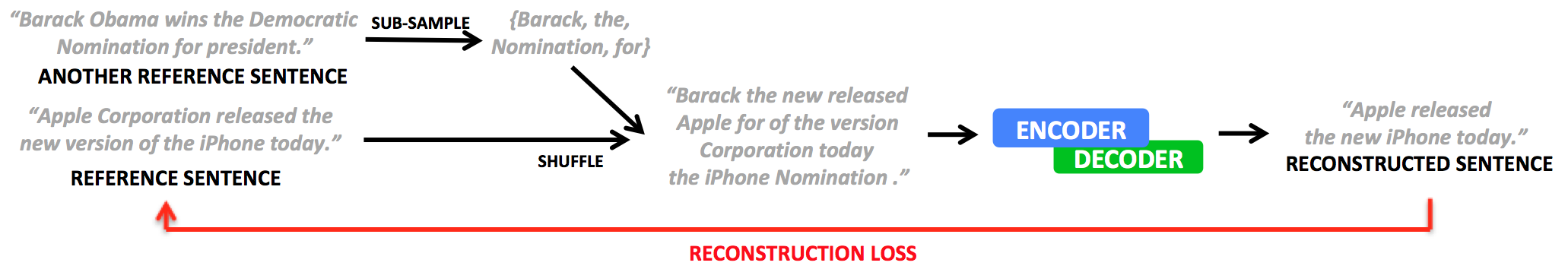}{Illustration of Additive Noising. A reference sentence is noised with subsampled words from another sentence, and then shuffled. The denoising auto-encoder is trained to recover the original reference sentence. This simulates a text summarization training regimes without the need for parallel corpora.}{0.35}{fig1}
    
    \paragraph{Additive Sampling} We randomly sample additional sentences from our data set, and then sub-sample a number of words from each without replacement. We then append the newly sampled words to our reference sentence. In our experiments, we sample two additional sentences for each reference sentence, and the number of words sampled from each is dependent on the length of the original reference sentence. In practice, we aim to generate a noised sentence that extends the original sentence by 40\% to 60\%. To fit the fully unsupervised learning paradigm, we do not introduce any biases into our sampling of words in training our model. In particular, we excluded approaches that overweighted adjectives or speaker identification (e.g ``said X on Tuesday") in noising.
    
    \paragraph{Shuffling} Next, we shuffle the resultant string of words. We experiment with two forms of shuffling: (i) a complete word (unigram) shuffle  and (ii) bigram shuffling, where we only shuffle among the word bigrams, keeping pairs of adjacent words together.
  
  This process is illustrated in Figure \ref{fig1}.
  
  \subsection{Length Countdown}
    To induce our model to output sequences of a desired length, we augment the RNN decoder in our model to take an additional \textit{length countdown} input. In the context of text generation, RNN decoders can be formulated as follows:
    \begin{equation}
    	\label{eq:length1}
    	h_{t} = \text{RNN}(h_{t-1}, x_t)
    \end{equation}
    where $h_{t-1}$ is the hidden state at the previous step and $x_t$ is an external input (often an embedding of the previously decoded token). Let $T_\text{dec}$ be the desired length of our output sequence. We modify (\ref{eq:length1}) with an additional input:
    \begin{equation}
    	\label{eq:length2}
    	h_{t} = \text{RNN}(h_{t-1}, x_t, T_\text{dec}-t)
    \end{equation}
    The length countdown $T-t$ is a single scalar input that ticks down to 0 when the decoder reaches the desired length $T$, and goes negative after. In practice, $(x_t, T_\text{dec}-t)$ are concatenated into a single vector. We also experimented with adding a length penalty to our objective function to amplify the loss from predicting the end-of-sequence token \texttt{<EOS>} at the desired time step, but did not find that our models required this additional loss term to output sequences of the desired length. 
    
  Explicit length control has been used in previous summarization work. \citet{fan2017controllable} introduced a length marker token that induces the model to target an output of a desired length, coarsely divided into discrete bins. \citet{Kikuchi} examined several methods of introducing target output length information, and found that they were effective without negatively impacting summarization quality. We found more success with our models with a per time-step input compared to a token at the start of the sequence as in \citet{fan2017controllable}.
    
  \subsection{Input Sentence Embedding}
    The model specified above is supplied only with an unordered set of words with which to construct a shorter sentence. However, there are typically many ways of ordering a given set of words into a grammatical sentence. In order to help our model better recover the original sentence, we also provide the model with an InferSent sentence embedding \cite{DBLP:journals/corr/ConneauKSBB17} of the original sentence, generated using a pre-trained InferSent model. The InferSent model is trained on NLI tasks, where, given a longer premise text and a shorter hypothesis text, the model is required to determine if the premise (i) entails, (ii) contradicts or (iii) is neutral to the hypothesis. The InferSent sentence embeddings are an intermediate output of the model, reflecting information captured from each text string. Conneau et al. show that InferSent sentence embeddings capture various aspects of the semantics of a string of text \cite{2018arXiv180501070C}, and should provide additional information to the model as to which ordering of words best match the meaning original sentence.
    
    We incorporate the InferSent embeddings by modifying the hidden state passed between the encoder and the decoder. In typical RNN encoder-decoder architectures, the final hidden state of the encoder is used as the initial hidden state of the decoder. In other words, $h_{0}^\text{dec}=h_{T_{\text{enc}}}^\text{enc}$. We learn a fully connected layer $f$ to be used as follows:	
    
    \begin{equation}
      h_{0}^\text{dec} = f(h_{T_\text{enc}}^\text{enc}, s)
    \end{equation}
where $s$ is the InferSent embedding of the input sentence. This transformation is only applied once on the hidden state shared from the encoder to the decoder. In the case of LSTMs, where there are both hidden states and cell states, we learn a fully connected mapping for each.

  \subsection{Numbered Out-of-Vocabulary (OOV) Embeddings}
    
    Many text summarization data sets are based on news articles and headlines, which often include names, proper nouns, and other rare words or tokens that may not appear in word embedding dictionaries. In addition, the output layer of most models are based on a softmax over all potential output tokens. This means that expanding the vocabulary to potentially include more rare words increases computation and memory costs in the final layer linearly. There are many approaches to tackle out-of-vocabulary (OOV) tokens \cite{see2017get,nallapati2016abstractive}, and we detail below our approach.
    
    To address the frequent occurrences of OOV characters, we learn a fixed number of embeddings for numbered OOV tokens.\footnote{We use a fixed number of 10 OOV tokens in our experiments.} Given an input sequence, we first parse the sentence to identify OOV tokens and number them in order, while storing the map from numbered OOV tokens to words.\footnote{In the case of shuffling and noising, we number the OOV tokens before shuffling, and number any additional OOV tokens from the noised input sentence in a second pass.} When embedding the respective tokens to be inputs to the RNN, we assign the corresponding embeddings for each numbered OOV token. We apply the same numbering system to the target, so the same word in the input and output will always be assigned the same numbered OOV token, and hence the same embedding. At inference, we replace any output numbered OOV tokens with their respective words. This allows us to output sentences using words not in our vocabulary. 
    
    This approach is similar to the pointer-generator model \cite{see2017get}, but whereas See et al. compute attention weights over all tokens in the input to learn where to copy and have an explicit switch between copying (pointer) and output (generator), we learn embeddings for a fixed number of OOV tokens, and the embeddings are in the same latent space as our pre-trained word embeddings.

\section{Experimental Setup}

	\subsection{Data} 
    For our text summarization task, We use the Annotated Gigaword \cite{Napoles:2012:AG:2391200.2391218} in line with \citet{rush2015neural}. This data set is derived for news articles, and consists of pairs of the main sentences in the article (longer), and the headline (shorter). The former and latter are used as references and summaries respectively in the context of summarization tasks. We preprocess the data using the scripts made available by the authors, which produces about 3.8M training examples and 400K validation examples. We sample randomly 10K examples for validation and 10K for testing from the validation set, similar to the procedure in \citet{nallapati2016abstractive}. Like \citet{rush2015neural}, we only extract the tokenized words of the first sentence, in contrast with \citet{nallapati2016abstractive} who extract the first two sentences as well as part-of-speech and named-entities tags.
    
    \subsection{Training} In training, we only use the reference sentences from the Gigaword dataset. For all our models, we used GloVe word embeddings \cite{glove}. We freeze these embeddings during training. Our vocabulary is comprised of the 20000 most frequent words in the references, and we use the aforementioned numbered OOV embeddings for other unseen words. We similarly freeze the InferSent model for sentence embeddings. The encoder and decoder are both 3-layer LSTMs with 512 hidden units. We use a batch size of 128, and optimize our models using Adam \cite{adam} with a initial learning rate of 0.0005, annealing it by 0.9 at every 10K mini-batches. We do not use dropout but use gradient clipping at 2. We train our models for 4 full epochs. 
    
    \subsection{Inference} At inference, we supply our model with the unmodified reference sentences--hence no noising is applied. We use the length countdown to target outputs of half the length of the reference sentences. The application of sentence embeddings is unchanged from training.

    \subsection{Implementation} We implemented our models using Pytorch \cite{paszke2017automatic}, and will make our code publicly available at \url{https://github.com/zphang/usc_dae}.

\section{Results}

	\subsection{ROUGE Evaluation} 
    
    In Table \ref{table:perf}, we evaluate our models on ROUGE \cite{Lin:2004} F1 scores, where a higher score is better. We provide a comparison with a simple but strong baseline, \textit{F8W} is simply first 8 words of the input, as is done in \citet{wang2018learning} and similarly to the \textit{Prefix} baseline (first 75 bytes) of \citet{rush2015neural}, as well as the ROUGE of the whole text with the target. We provide scores of two supervised text-summarization methods on Gigaword. One is our own baseline, consisting of a sequence-to-sequence attentional encoder-decoder trained on pairs of reference and target summary text, but incorporating the same length countdown mechanism as in our unsupervised models. The other is the \textit{words-lvt2k-1sent} model of \citet{nallapati2016abstractive}. Although not their best model, it is most comparable to ours since it only uses the first sentence and does not extract \textit{tf-idf} vectors nor named entities tags. 
    
    \textit{F8W} and \textit{All text} are strong baselines due to the tendency of news articles to contain specific terms that are rarely rephrased. We find that our models perform competitively with these baselines, although they pale in comparison to supervised methods, likely because they do not learn any style transfer and use only the reference's vocabulary and writing style. While our ROUGE-1 scores are in line with the baselines, our ROUGE-2 scores fall somewhat behind. Including InferSent sentence embeddings improves our ROUGE scores across the board. Our supervised baseline performance is close to that of  \citet{nallapati2016abstractive}, with results lower in ROUGE-2 likely due to their use of beam search. Nevertheless, the supervised baseline is representative of the performance of a standard sequence-to-sequence attentional model on this task.
    
    A direct comparison of ROUGE scores is not completely adequate for evaluating our model. Because of our training regime, our model primarily learned to generate shortened sentences that often still retain the style of the input sentences. Unlike other model setups, our model has never been exposed to any examples of summaries, and hence never adapts its output to match the style of the target summaries. In the case of Gigaword, the summaries are headlines from news articles, which are written in a particular linguistic style (e.g. dropping articles, having clauses rather than full sentences). ROUGE will thus penalize our model, that tends to output longer, full sentences. In addition, ROUGE is an imperfect metric for summarization as word/$n$-gram overlap does not fully capture summary relevancy and retention of meaning.\footnote{See discussion in \citet{nallapati2016abstractive}, or in \citet{paulus2017deep} where a reinforcement learning model trained on a Rouge-L objective alone achieves the best scores but ``produces the least readable summaries among [their] experiments"} For this reason, we also conduct a separate human evaluation of our different models against a supervised baseline (Section \ref{humaneval}).

\begin{table*}
\renewcommand{\arraystretch}{1.2}
\centering
\small
\begin{tabular}{ l c  c  c  c}
\toprule
 & \multicolumn{3}{c}{\textbf{ROUGE}} & \\ \cmidrule{2-4}
\textbf{Model} & \textbf{R-1} & \textbf{R-2} & \textbf{R-L} & \textbf{Avg. Length} \\ \midrule
\textit{Baselines:} & \multicolumn{4}{c}{ }\\ 
\ \ All text & \textbf{28.91} & \textbf{10.22} & \textbf{25.08} & 31.3 \\
\ \ F8W & 26.90 & 9.65 & 25.19 & 8 \\ 

\textit{Unsupervised (Ours):} & \multicolumn{4}{c}{ }\\ 

\ \ 2-g shuf & 27.72 & 7.55 & 23.43 & 15.4 \\ 

\ \ 2-g shuf + InferSent & \textbf{28.42} & \textbf{7.82} & \textbf{24.95} & 15.6 \\ 

\textit{Supervised abstractive:} & \multicolumn{4}{c}{ }\\ 
\ \ Seq2seq & \textbf{35.50} & 15.54 & 32.45 & 15.4 \\ 

\ \ (words-lvt2k-1sent) \cite{nallapati2016abstractive}  & 34.97 & \textbf{17.17} & \textbf{32.70} & - \\ 
\bottomrule
\end{tabular}
\caption{Performance of Baseline, Unsupervised and Supervised Models. 
Our unsupervised models pale in comparison to supervised models, and perform in line with baselines. Simple baselines in text summarization benchmarks tend to be unusually strong. The unsupervised model incorporating sentence embeddings performs slightly better on ROUGE.
}
\label{table:perf}
\end{table*}

\begin{figure}[t]
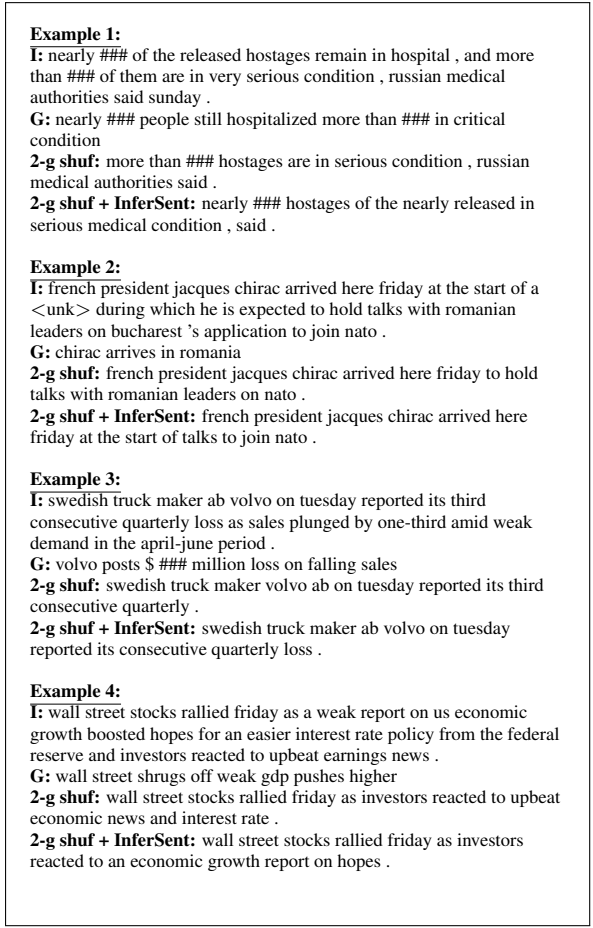

    \scriptsize
    \begin{framed}    
\begin{flushleft}

  \textbf{\underline{Example 1:}}\\
  \textbf{I:}  nearly \#\#\# of the released hostages remain in hospital , and more than \#\#\# of them are in very serious condition , russian medical authorities said sunday . \\
  \textbf{G:} nearly \#\#\# people still hospitalized more than \#\#\# in critical condition \\
  \textbf{2-g shuf:} more than \#\#\# hostages are in serious condition , russian medical authorities said .\\
  \textbf{2-g shuf + InferSent:} nearly \#\#\# hostages of the nearly released in serious medical condition , said . \\
\ \\

  \textbf{\underline{Example 2:}}\\
  \textbf{I:}  french president jacques chirac arrived here friday at the start of a $<$unk$>$ during which he is expected to hold talks with romanian leaders on bucharest 's application to join nato . \\
  \textbf{G:} chirac arrives in romania \\
  \textbf{2-g shuf:} french president jacques chirac arrived here friday to hold talks with romanian leaders on nato .\\
  \textbf{2-g shuf + InferSent:} french president jacques chirac arrived here friday at the start of talks to join nato .
 \\
\ \\

  \textbf{\underline{Example 3:}}\\
  \textbf{I:} swedish truck maker ab volvo on tuesday reported its third consecutive quarterly loss as sales plunged by one-third amid weak demand in the april-june period . \\
  \textbf{G:} volvo posts \$ \#\#\# million loss on falling sales \\
  \textbf{2-g shuf:} swedish truck maker volvo ab on tuesday reported its third consecutive quarterly .
\\
  \textbf{2-g shuf + InferSent:} swedish truck maker ab volvo on tuesday reported its consecutive quarterly loss .
 \\
\ \\

  \textbf{\underline{Example 4:}}\\
  \textbf{I:} wall street stocks rallied friday as a weak report on us economic growth boosted hopes for an easier interest rate policy from the federal reserve and investors reacted to upbeat earnings news .
 \\
  \textbf{G:} wall street shrugs off weak gdp pushes higher\\
  \textbf{2-g shuf:} wall street stocks rallied friday as investors reacted to upbeat economic news and interest rate .\\
  \textbf{2-g shuf + InferSent:} wall street stocks rallied friday as investors reacted to an economic growth report on hopes . \\
\ \\

\end{flushleft}
    \end{framed}
    \caption{\footnotesize \label{fig:examples} Examples of inputs, ground-truth summaries, and outputs from two of our models. \textbf{I} is input, \textbf{G} (gold) is the true summaries. 
    Example 1 and 2 show our models summarizing pertinent information from the input. 
    Example 3 demonstrates the ability to recover long ordered strings of tokens, even though the models are trained on shuffle data.
    Example 4 shows cases where the models output grammatical but semantically incorrect sentences.}
    \label{fig:examples}
  \end{figure}

\subsection{ROUGE Ablation study}

In Table \ref{table:ablation}, we report the results  of an ablation study. We observe that all three components we vary, namely the use of attention, bigram shuffling, and incorporation of sentence embeddings, contribute positively to the performance of our model as measured by ROUGE. The model that incorporates all three obtains the highest ROUGE scores.

\begin{table}
\renewcommand{\arraystretch}{1.2}
\centering
\small
\begin{tabular}{  l  c  c  c }
\toprule
 & \multicolumn{3}{c}{\textbf{ROUGE}} \\ \cmidrule{2-4}
\textbf{Model} & \textbf{R-1} & \textbf{R-2} & \textbf{R-L} \\ \midrule
1-g shuf (w/o attn) & 23.01 & 5.51 & 20.07 \\
2-g shuf (w/o attn) & 22.36 & 5.18 & 19.60 \\
1-g shuf & 27.22 & 7.63 & 23.55  \\
2-g shuf & 27.72 & 7.55 & 23.43 \\
1-g shuf + InferSent & 28.12 & 7.75 & 24.81  \\
2-g shuf + InferSent & \textbf{28.42} & \textbf{7.82} & \textbf{24.95} \\
  
  \bottomrule
  \end{tabular}
\caption{Ablation study. We find that using attention, shuffling bigrams, and incorporating sentence embeddings all improve our ROUGE scores. All length countdowns settings are the same is in the main model.}
\label{table:ablation}
\end{table}

\subsection{Impact of Length}

To assess our models' ability to deal with sequences of text of different length, we measure the ROUGE scores on two bins of length of the input text, from 16 to 30 tokens and from 31 to 45. As expected, longer sentences pose a harder challenge to the model, with our model performing better on shorter than longer sentences. Across most sequence-based problems, models tend to perform better on shorter sequences. However, in the context of the text summarization or sentence compression, longer sentences not only contain more information that the model would need to selective remove, but also more information from which to identify the central theme of the sentence.

\subsection{Human Evaluation} \label{humaneval}
	To qualitatively evaluate our model, we take inspiration from the methodology of \citet{turner2005supervised} to design our human evaluation. We asked 6 native English speakers to evaluate randomly chosen summaries from five models: our best models with and without InferSent sentence embeddings, a summary generated from a trained supervised model, and the ground truth summary. The sentences are evaluated based on two separate criteria: the grammaticality of the summary and how well it retained the information of the original sentence. In the former, only the summary is provided, whereas in the latter, the evaluator is shown both the original sentence as well as the summary. Each of these criteria were graded on a scale from 1 to 5. The examples are from the test set, with 50 examples randomly sampled for each evaluator and criterion.\footnote{The sampling is constrained to ensure each evaluator sees an equal number of summaries from each model, although evaluators are informed neither about the sampling process, nor how many or what models are involved.} 
    
We report the average evaluation given by our 6 evaluators in Table \ref{table:human}. That the \emph{Meaning} score for the ground truth is somewhat low (3.87) is not surprising. Within the Gigaword dataset, summaries (headlines) sometimes include information not within the reference (main line of the article). We observe that quantitative evaluation does not correlate well with human evaluation. Methods using InferSent embeddings improved our ROUGE scores but perform worse in human evaluation, which is in line with the summaries presented in \ref{fig:examples}. Notably, the model trained on shuffled bigrams and InferSent embeddings performed best within our ablation study, but the worst among the three models in human evaluation. Encouragingly, the model without InferSent embeddings performs competitively with the supervised baseline in both grammar and meaning scores, indicating that although it does not capture the style of headlines, it succeeds in generating grammatical sentences that roughly match the meaning in the reference. Some evaluators highlighted that it was problematic to rate meaning for ungrammatical sentences.

\begin{table}
\renewcommand{\arraystretch}{1.2}
\centering
\small
\begin{tabular}{ l c c c c }
\toprule
& \multicolumn{3}{c}{\textbf{ROUGE}} &  \\ \cmidrule{2-4}
\textbf{Input Length} & \textbf{R-1} & \textbf{R-2} & \textbf{R-L} & \textbf{Avg. Length}\\ \midrule
16-30 & 30.79 & 9.20 & 27.73 & 12.6\\
31-45 & 26.89 & 6.76 & 23.04 & 17.7\\
\bottomrule
\end{tabular}
\caption{Effect of input sentence length on performance, using the 2-g shuf + InferSent model. Performance tends to be worse on longer input texts.}
\label{table:length}
\end{table}

\begin{table}
\renewcommand{\arraystretch}{1.2}
\centering
\small
\begin{tabular}{  l  c  c }
\toprule
\textbf{Model} & \textbf{Grammar} & \textbf{Meaning}  \\ \midrule
2-g shuf              &  \textbf{3.53 ($\pm$0.18)} &  2.53 ($\pm$0.16) \\
1-g shuf + InferSent  &  2.82 ($\pm$0.17) &  2.50 ($\pm$0.15) \\
2-g shuf + InferSent  &  2.87 ($\pm$0.16) &  2.13 ($\pm$0.13) \\
Seq2seq (Supervised)  &  3.43 ($\pm$0.18) &  \textbf{2.60 ($\pm$0.17)} \\ \midrule
Ground Truth                 &  4.07 ($\pm$0.13) &  3.87 ($\pm$0.16) \\
  \bottomrule
  \end{tabular}
\caption{Human Evaluation. Mean scores, with 1 standard error confidence bands in parentheses. 
Our best model performs competitively with a supervised baseline in both grammatical correctness and retuention of meaning.
Models with sentence embeddings perform worse in human evaluation, despite obtaining better ROUGE scores.
}
\label{table:human}
\end{table}

\subsection{Output Analysis}

We show in Figure \ref{fig:examples} several examples of the inputs, ground-truths target summaries, and outputs from 2 of our models. We observe that the output sentences are generally well-conditioned though occasionally imperfectly grammatical. We also observe certain artifacts from training only on reference texts that are not reflected in ground-truth summaries. For example, every output sentence ends with a period, and several examples end with speaker identification clauses. In all instances, we observe that the model without InferSent outputs sentences are more readable and relevant, confirming human evaluation results in \ref{table:human}. 

Example 1 shows that our model can extract the most pertinent information to generate a grammatical summary that captures the original meaning.

Example 2 shows an instance where our output accurately summarizes the input text despite low ROUGE scores to the target (R-1 of 21.1 and R-2 of 11.8). In this case, both models capture the core meaning of the input.

Example 3 shows that although the models are provided completely shuffled words in training, at inference it is able to recover complex terms such as ``swedish truck maker ab volvo". We note that this may be a bias in the data set (sentences in news often start with proper nouns preceded by qualifiers) and hence a simple strategy for the model to discover. This examples also shows common mistake of our models: in the output of model without InferSent, it drops an important word (``loss") right before the end of the sentence, causing it to fail to capture the original meaning.

Example 4 shows that on longer sentences, our models may sometimes fail to accurately capture meaning. In this case, for the model without InferSent, although the output is grammatical and meaningful, it captures a meaning different than that of the original input. Indeed, our model suggests that upbeat news cause the rally whereas the original sentence indicates that given poor economic news investors anticipate easier monetary policy and thus caused a stock rally. 

\subsection{Length Variation}

Because the desired length of the output sequence is a user-defined input in the model, we can take an arbitrary sentence and use the model to output the corresponding compressed (or even expanded) sentence of any desired length. We show two examples in Figure \ref{fig:length}, where we vary the desired length from 7 to the input length, using our best model based on human evaluation. We observe that for very short desired lengths, the model struggles to produce meaningful sentences, whereas for desired lengths close to the input length, the model nearly reconstructs the input sentence. Nevertheless, we observe that for many of the intermediate lengths, the model outputs sentences that are close in meaning to the input sentence, with different ways of rephrasing or shortening the input sentence in the interim. This suggests that when the ratio of the desired output sentence length to the input sentence length is close to that of the training regime, the model is able to perform better than when it has to generate sentences with other ratios.

\begin{figure}[t]
    \scriptsize
    \begin{framed}    
\begin{flushleft}
  \textbf{\underline{Example 1:}}\\
  \textbf{I:} three convicted serial killers have been hanged in tehran 's evin prison , the khorasan newspaper reported sunday .\\
\textbf{L=9:} three convicted serial killers have been hanged in . \\
\textbf{L=11:} three convicted serial killers have been hanged in prison sunday . \\
\textbf{L=13:} three convicted serial killers have been hanged , a newspaper reported sunday . \\
\textbf{L=15:} three convicted serial killers have been hanged in tehran , a newspaper reported sunday . \\
\textbf{L=17:} three convicted serial killers have been hanged in tehran , the tehran 's newspaper reported sunday . \\
\textbf{L=19:} three convicted serial killers have been hanged in tehran 's prison , the newspaper tehran newspaper reported sunday . \\
\ \\
  \textbf{\underline{Example 2:}}\\
  \textbf{I:} a home-made bomb was found near a shopping center on indonesia 's ambon island , where \#\# people were wounded by an explosion at the weekend , state media said on monday . \\
\textbf{L=9:} a home-made bomb explosion wounded \#\# people monday . \\
\textbf{L=11:} a home-made bomb explosion wounded \#\# people on indonesia monday . \\
\textbf{L=13:} a home-made bomb explosion wounded \#\# people on indonesia 's ambon island . \\
\textbf{L=15:} a home-made bomb explosion wounded \#\# people at a shopping center on ambon monday . \\
\textbf{L=17:} a home-made bomb explosion wounded \#\# people at a shopping center on ambon island on monday . \\
\textbf{L=19:} a home-made bomb explosion wounded \#\# people at a shopping center near ambon on indonesia 's island state . \\
\textbf{L=21:} a home-made bomb explosion wounded \#\# people at a shopping center near ambon on indonesia 's ambon island on monday . \\
\textbf{L=23:} a home-made bomb was found on a shopping center near ambon , indonesia 's state on monday , state media said monday . \\
\textbf{L=25:} a home-made bomb was found on a shopping center near ambon , indonesia 's state media center where \#\# people were wounded by bomb . \\
\textbf{L=27:} a home-made bomb was found on a shopping center near ambon , indonesia 's state media center where \#\# people were wounded , media said monday . \\
\textbf{L=29:} a home-made bomb was found on a shopping center near ambon , indonesia 's state media on monday , where \#\# people were wounded by an explosion nearby . \\
\textbf{L=31:} a home-made bomb was found on a shopping center near ambon , indonesia 's state media on monday , where \#\# people were wounded by an explosion at the weekend . \\
\textbf{L=33:} a home-made bomb was found on a shopping center near ambon , indonesia 's state media on monday , where \#\# people were wounded by an explosion at the weekend on monday . \\
\end{flushleft}
    \end{framed}
    \caption{\footnotesize \label{fig:length} Summaries of varied desired lengths, using the 2-g shuf model. \textbf{L} is the desired output length provided to the model. Because the desired output length is a human-provided input, we can produce summaries of varying lengths, ranging from highly contracted to verbose.} 
    \label{fig:length}
  \end{figure}

\section{Discussion}

In our experiments, we found that denoising auto-encoders quickly learn to generate well-conditioned text, even from badly conditioned inputs. We were surprised by the ability of denoising auto-encoders to recover readable sentences even from completely shuffled and noised sets of words. We observed some cases where the denoising auto-encoders outputs sequences that are grammatical correct but nonsensical or semantically different from the input. However, the ability for denoising auto-encoders to subsample words to form grammatical sentences would significantly reduce the search space for candidate sentences, and we believe this could be useful for tasks involving sentence construction and reformulation.

Our attempts to better condition the denoising auto-encoders outputs on the original sentence using sentence embeddings had mixed results. Although the incorporation of InferSent embeddings improved our quantitative ROUGE scores, human evaluators scored outputs conditioned on InferSent embeddings markedly worse on both grammar and meaning retention. It is unclear whether this is due to InferSent embeddings failing to capture the most significant semantic information, or if our mechanism for incorporating the sentence embedding is suboptimal. 

Lastly, we echo sentiments from previous authors that ROUGE remains an imperfect proxy for measuring the adequacy of summaries. We found that ROUGE scores can be fairly uncorrelated with human evaluation, and in general can be distorted by quirks of the data set or model outputs, particularly pertaining to length, formatting, and handling of special tokens. On the other hand, human evaluation can be more sensitive to comprehensibility and relevancy while being more robust to rewording and reasonable ambiguity. Based on our human evaluation, we find that both unsupervised and supervised methods still fall short of effective sentence compression and summarization.

\section{Conclusion}

We present a fully unsupervised approach to the task of sentence compression in the form of a denoising auto-encoder with additive noising and word shuffling. Our model achieves comparable scores in human evaluation to a supervised sequence-to-sequence attentional baseline in grammatical correctness and retention of meaning, but underperforms on ROUGE. Output analysis indicates that our model does not capture the particular style of the summaries in the Gigaword dataset, but nevertheless produces reasonably valid sentences that capture the meaning of the input. Although our models are still prone to making mistakes, they provide a strong baseline for future sentence compression and summarization work.

\section*{Acknowledgments}

We would like to express our gratitude to Sam Bowman for his thoughtful advice and feedback in the writing of this paper.
We thank the NVIDIA Corporation for their support.

\bibliography{conll2018}
\bibliographystyle{acl_natbib_nourl}

\end{document}